\documentclass[fleqn,10pt]{wlscirep}
\usepackage{multirow}
\usepackage[normalem]{ulem}
\usepackage[caption=false]{subfig}
\useunder{\uline}{\ul}{}
\usepackage{lineno}
\usepackage{afterpage}

\title{Multimodal Machine Learning-based Knee Osteoarthritis Progression Prediction from Plain Radiographs and Clinical Data}

\author[1,8,*]{Aleksei Tiulpin}
\author[2]{Stefan Klein}
\author[3,4]{Sita M.A. Bierma-Zeinstra}
\author[1]{J\'er\^ome Thevenot}
\author[5]{Esa Rahtu}
\author[6]{Joyce van Meurs}
\author[7]{Edwin H.G. Oei}
\author[1,8]{Simo Saarakkala}

\affil[1]{Research Unit of Medical Imaging, Physics and Technology, University of Oulu, Oulu, Finland.}
\affil[2]{Biomedical Imaging Group Rotterdam, Depts. of Medical Informatics \& Radiology, Erasmus MC, University Medical Center Rotterdam, the Netherlands.}
\affil[3]{Department of General Practice, Erasmus MC, University Medical Center Rotterdam, the Netherlands}
\affil[4]{Department of Orthopedics, Erasmus MC, University Medical Center Rotterdam, the Netherlands.}
\affil[5]{Department of Signal Processing, Tampere University of Technology, Tampere, Finland.}
\affil[6]{Department of Internal Medicine, Erasmus MC, University Medical Center Rotterdam, the Netherlands}
\affil[7]{Department of Radiology \& Nuclear Medicine, University Medical Center Rotterdam, the Netherlands}
\affil[8]{Department of Diagnostic Radiology, Oulu University Hospital, Oulu, Finland}

\affil[*]{aleksei.tiulpin@oulu.fi}


\begin{abstract}
Knee osteoarthritis (OA) is the most common musculoskeletal disease without a cure, and current treatment options are limited to symptomatic relief. Prediction of OA progression is a very challenging and timely issue, and it could, if resolved, accelerate the disease modifying drug development and ultimately help to prevent millions of total joint replacement surgeries performed annually. Here, we present a multi-modal machine learning-based OA progression prediction model that utilizes raw radiographic data, clinical examination results and previous medical history of the patient. We validated this approach on an independent test set of 3,918 knee images from 2,129 subjects. Our method yielded area under the ROC curve (AUC) of 0.79 (0.78-0.81) and Average Precision (AP) of 0.68 (0.66-0.70). In contrast, a reference approach, based on logistic regression, yielded AUC of 0.75 (0.74-0.77) and AP of 0.62 (0.60-0.64). The proposed method could significantly improve the subject selection process for OA drug-development trials and help the development of personalized therapeutic plans.
\end{abstract}

\newcommand{\rpm}{\raisebox{.2ex}{$\scriptstyle\pm$}}
\newcommand\T{\rule{0pt}{2.6ex}}       %
\newcommand\B{\rule[-1.2ex]{0pt}{0pt}} %

\begin{document}
\flushbottom
\maketitle

\thispagestyle{empty}

\section*{Introduction}
Knee osteoarthritis (OA) is the most common musculoskeletal disorder causing significant disability for patients worldwide \cite{arden2006osteoarthritis}. OA is a degenerative disease and there is a lack of knowledge on the factors contributing to its progression. The overall etiology of OA is also not understood and there is no effective treatment, besides behavioral interventions. Furthermore, at the end stage of the disease, the only available treatment option is total knee replacement (TKR) surgery, which is highly invasive, costly and also strongly affects the patient’s quality of life. OA is a major burden for the public health care system and it is increasing further with the aging of the population. For example, according to the statistics only in the United States, around 12\% of the population suffer from OA and the annual rate of TKR for people 45-64 years of age has doubled since the year of 2000 \cite{ferket2017impact}. From the economical point of view, OA causes enormous costs for society and the costs of these surgeries are estimated to be over nine billion euros \cite{ferket2017impact}.

In primary health care, OA is currently diagnosed based on a combination of clinical history, physical examination, and X-ray imaging (radiography) if needed. However, the current widely available diagnostic modalities do not allow for effective OA prognosis assessment \cite{bedson2005prevalence}, which is important for the planning of appropriate therapeutic interventions and also for recruitment to OA disease modifying drugs development trials \cite{jamshidi2018machine}. A possible improvement would be to extend this diagnostic chain with Magnetic Resonance Imaging (MRI), which is, however, costly, time-consuming, has limited availability and not applicable for wide use \cite{van2018general}.

While being imperfect and lacking decision consistency, the current OA diagnostic tools can be enhanced using computer-assisted methods. For example, it has been shown that the gold clinical standard for OA severity assessment from radiographs, semi-quantitative Kellgren-Lawrence (KL) \cite{kellgren1957radiological} system that highly suffers from subjectivity of a practitioner, can be automated using Deep Learning  -- a state-of-the-art Machine Learning approach widely used in computer vision \cite{tiulpin2018automatic,norman2018applying,antony2016quantifying}. However, to the best of our knowledge, there have been no similar studies on Deep Learning-based prediction of structural knee OA progression, in which the raw image data are directly used for prediction instead of the KL grades defined by a radiologist.

Current state-of-the-art OA progression prediction models are based on a combination of texture descriptors that are calculated from imaging, KL-grade, clinical and anthropometric data \cite{kerkhof2014prediction,janvier2017subchondral,janvier2017subchondral2,kraus2009trabecular}. However, their performance and generalizability are difficult to assess for multiple reasons. Firstly, the texture descriptors may suffer from sensitivity to data acquisition settings. This can lead to limited sample size, as is for example seen in the studies of Janvier \textit{et al.}, where only non-processed digital images were used \cite{janvier2017subchondral, janvier2017subchondral2}. Secondly, only a few current progression studies used an external dataset besides the one that was utilized to develop the prediction model \cite{kerkhof2014prediction,yu2019development,hosnijeh2018development}. If such external dataset is not utilized, this can lead to a possible overfitting and eventually a bias in the final results \cite{tiulpin2018automatic}. Finally, it has been previously shown that most of the OA evolution modelling studies tend to focus on estimating the decrease of joint space width (JSW) as a measure of progression \cite{emrani2008joint}. Such outcome can be challenging to validate due to inherent problems associated with radiographic data acquisition (e.g. varying beam angle) and it does not depict all the changes happening with the joint. Previously, it has been recommended to assess OA progression using measures that incorporate the information about both -- JSW and osteophytes \cite{lavalley2001validity}, \textit{i.e.}, treating future increase of the KL-grade as a progression outcome.

\begin{figure}[hbt!]
\centering
\includegraphics[width=\textwidth]{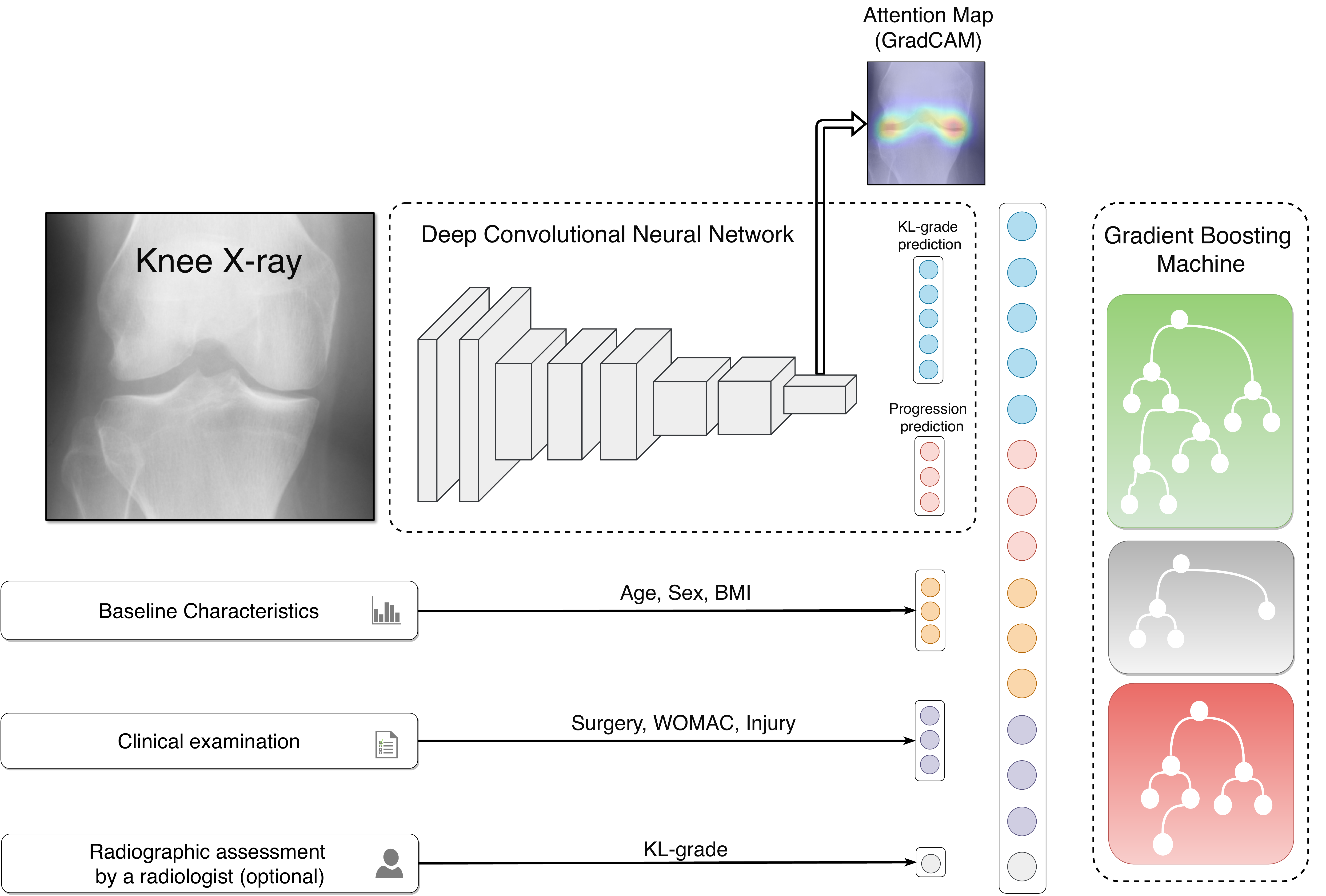}
\caption{Schematic representation of our multi-modal pipeline, predicting the risk of osteoarthritis (OA) progression for a particular knee. We first use a Deep Convolutional Neural Network (CNN), trained in a multi-task setting to predict the probability of OA progression (no progression, rapid progression, slow progression) and the current stage of OA defined according to the Kellgren-Lawrence (KL) scale.  Subsequently, we fuse these predictions with patient’s Age, Sex, Body-Mass Index, given knee injury and surgery history, symptomatic assessment results and, optionally, a KL grade given by a radiologist using a Gradient Boosting Machine Classifier. After obtaining prediction from CNN, we utilize GradCAM attention maps to make our method more transparent and highlight the zones in the input knee radiograph, which were considered most important by the network.}\label{img:workflow}
\end{figure}

In this study, we propose a novel method based on Machine Learning that directly utilizes raw radiographic data, physical examination, patient’s medical history, anthropometric data and, optionally, a radiologist’s statement (KL-grade) to predict structural OA progression. Here, we aim to predict any increase of a current KL-grade or potential need for TKR within the next 7 years after the baseline examination for patients having no, early or moderate OA. Our method employs a Deep Convolutional Neural Network (CNN) \cite{schmidhuber2015,lecun2015deep} that evaluates the probability of OA progression jointly with the current OA severity in the analyzed knee as an auxiliary outcome. Further, we improve the prognosis from CNN by fusing its prediction with the clinical data using a Gradient Boosting Machine (GBM) \cite{friedman2001greedy}. Schematically, our method is presented in Figure \ref{img:workflow}. 

\section*{Results}
\subsection*{Training and testing datasets}
We used the metadata provided in Osteoarthritis Initiative (OAI) and Multicenter Osteoarthritis Study (MOST) cohorts to select progressors and non-progressors for train and test datasets, respectively. We considered only the knees having no, early or moderate OA (KL-0, KL-1, KL-2 and KL-3) at the baseline (first visit) as these are the most relevant clinical cases. Furthermore, we excluded from the test set all the subjects who died between the follow-ups for coherence of our data. Additionally, the subjects who did not progress and dropped out from the study before the last follow-up examination were excluded. 
After the pre-selection process, we used 4,928 knees (2,711 subjects) from OAI dataset for training and 3,918 knees (2,129 subjects) from MOST dataset for testing of our model. Here, 1,331 (27\%) and 1,501 (47\%) knees were identified as progressors in OAI and MOST data, respectively. As a progression definition, we utilized an increase of a KL-grade within the following years. Here, we ignored the increase from KL-0 to KL-1 and included all cases with progression to TKR.  To harmonize the data between OAI and MOST datasets, we defined the following three fine-grained categories:
\begin{itemize}
    \item $y=0$: no knee OA progression
    \item $y=1$: progression within the next 60 months (fast progression)
    \item $y=2$: progression after 60 months (slow progression)
\end{itemize}

Supplementary Tables \ref{supp_tab:subjects} and \ref{supp_tab:knees} describe the training and the testing sets derived from OAI and MOST datasets respectively.

\subsection*{Reference methods}
\begin{figure}[htbp]
\centering

 \subfloat[]{%
   \includegraphics[width=0.48\textwidth]{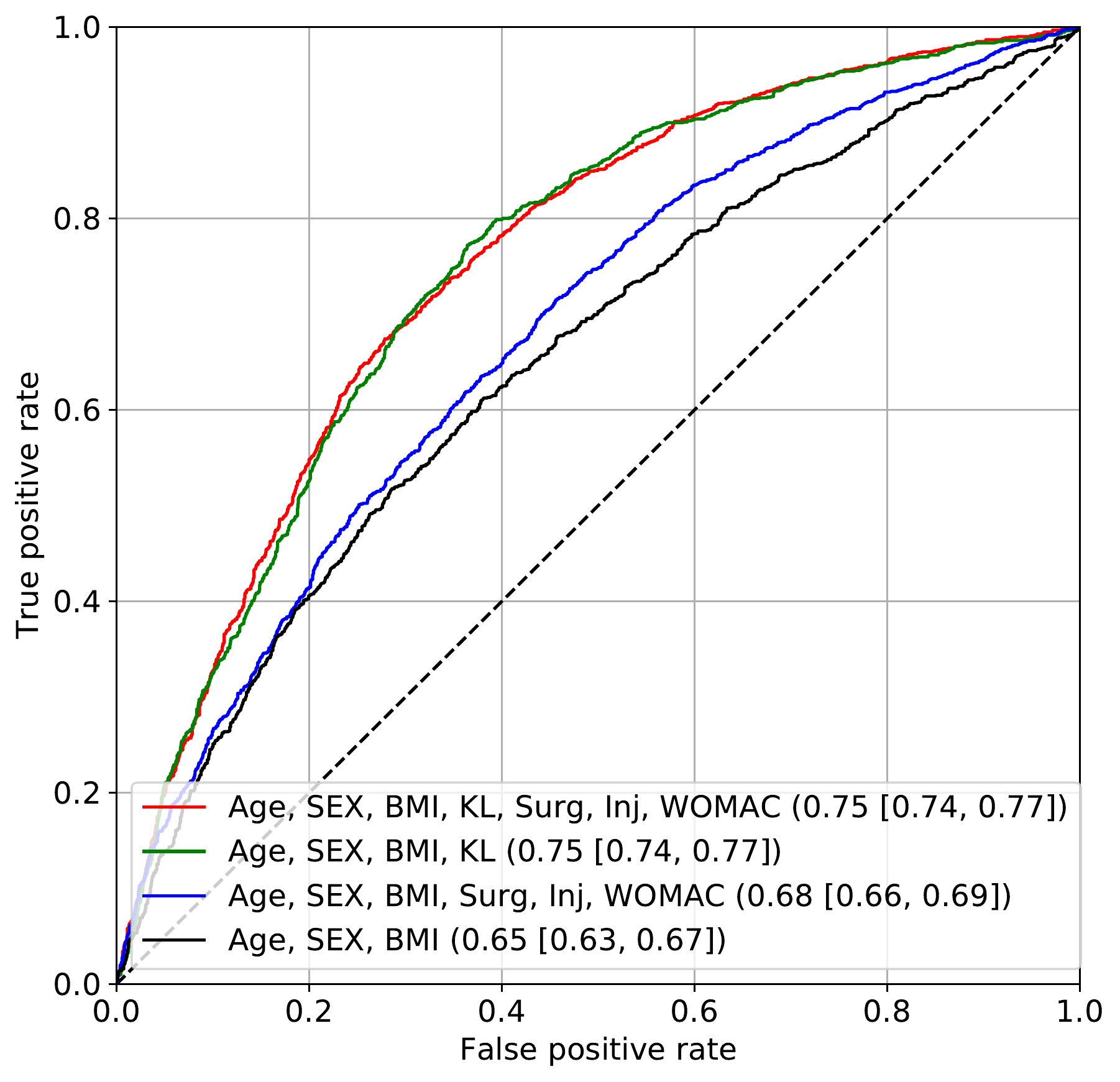}
   }
   \hfill
 \subfloat[]{%
   \includegraphics[width=0.48\textwidth]{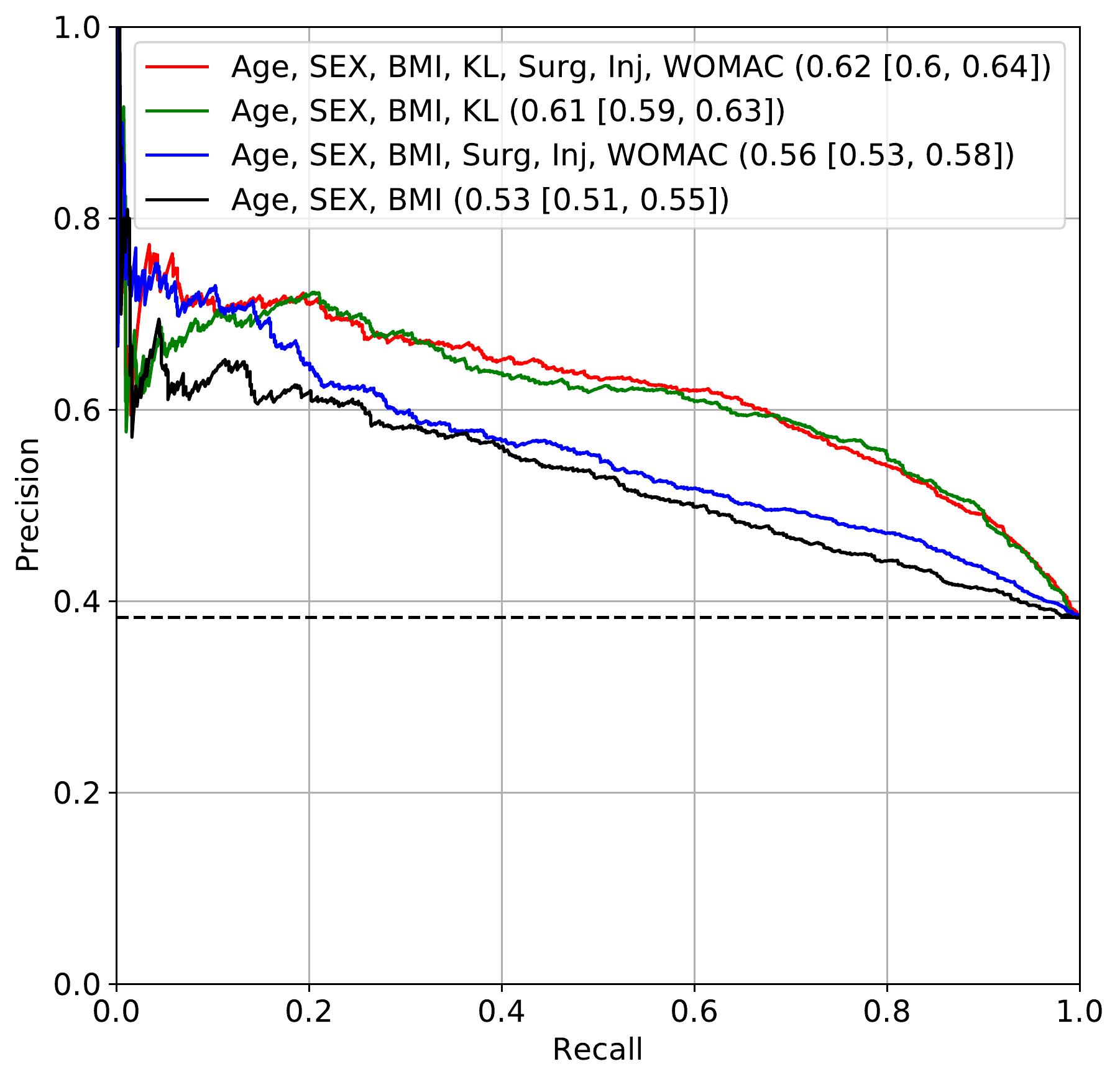}
   }
   
\caption{Assessment of Logistic Regression-based models’ performance. The subplot (a) demonstrates the ROC curves and the subplot (b) precision-recall curves. Black dashed lines indicate the performance of a random classifier in case of AUC, and performance of the prediction model based on the dataset labels distribution. The subplots’ legends reflect the benchmarked models and the values of corresponding metrics with 95\% confidence intervals. Here, Area under the ROC curve metric is used in subplot (a) and Average Precision in subplot (b).}\label{img:lr_bench}
\end{figure}

Firstly, we utilized several reference methods (see details in Methods) in order to understand the added value of our approach. These models were trained to predict a probability $P(y>0 | x)$ of a particular knee $x$ to have a KL-grade increase in the future. Here, we pooled the classes $y=1$ and $y=2$ together to derive a binary outcome, which was used in both Logistic Regression (LR) and GBM reference methods.
In Figure \ref{img:lr_bench}, we demonstrate the performance of LR, which is commonly used in OA research \cite{hosnijeh2018development,yu2019development,janvier2017subchondral,kerkhof2014prediction}. All of the LR models were derived and tested on the existing image assessment and clinical data provided by the OAI and MOST datasets,  respectively. In cross-validation experiments on OAI data, we also assessed the added value of regularization \cite{friedman2001elements} and found no difference between regularized and non-regularized LR models.

From Figure \ref{img:lr_bench}, it can be seen that two best models exist: one based on Age, Sex, Body-Mass Index and KL grade (model 1), and the other being the same with the addition of symptomatic assessment (Western Ontario and McMaster Universities Arthritis Index, WOMAC \cite{bellamy1988validation}), injury and surgery history (model 2). We chose the latter in our further comparisons because it performs with higher precision at lower recall while yielding similar performance at other recall levels. This model yielded AUC of 0.75 (0.74-0.77) and Average Precision (AP) of 0.62 (0.60-0.64). All the mentioned risk factors included into the reference models were selected on the basis of their use in the previous studies \cite{yu2019development,kerkhof2014prediction,hosnijeh2018development}.

\begin{figure}[ht!]
 \centering
 \subfloat[]{%
   \includegraphics[width=0.48\textwidth]{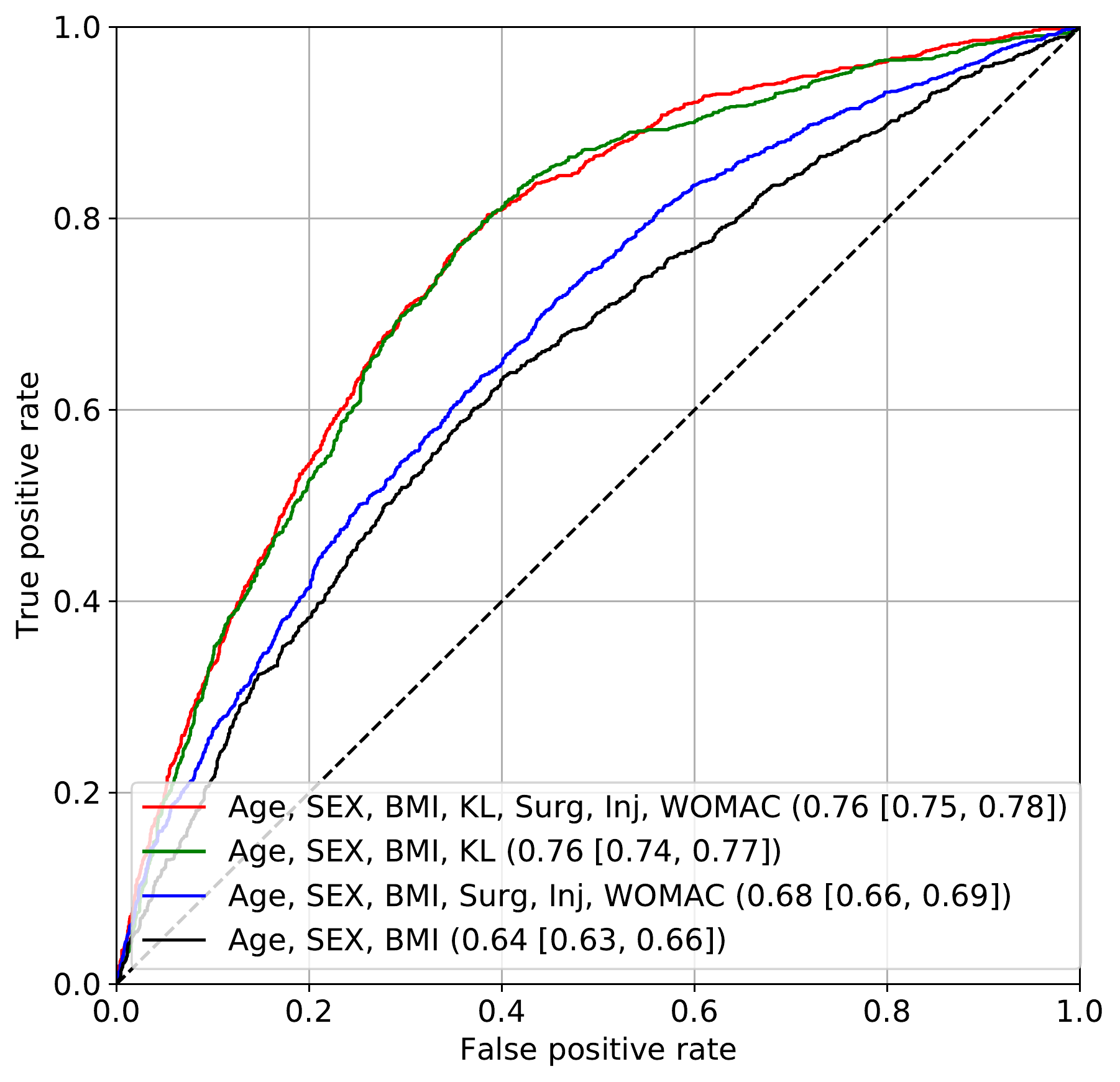}
   }
   \hfill
 \subfloat[]{%
   \includegraphics[width=0.48\textwidth]{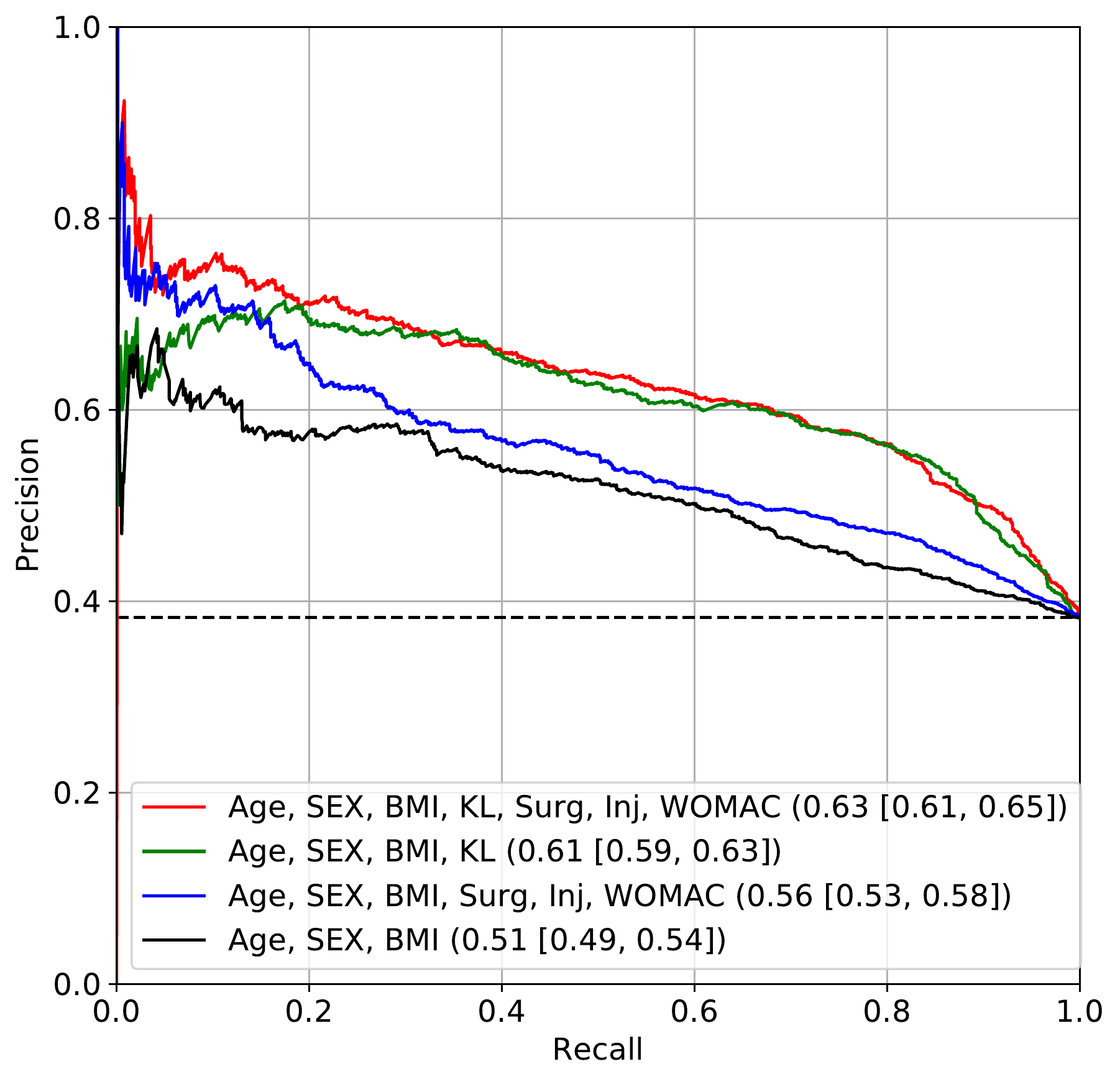}
   }
   
 \caption{Assessment of Gradient Boosting Machine-based models’ performance. The subplot (a) demonstrates the ROC curves and the subplot (b) precision-recall curves. Black dashed lines indicate the performance of a random classifier in case of AUC, and performance of the prediction model based on the dataset labels distribution. The subplots’ legends reflect the benchmarked models and the values of corresponding metrics with 95\% confidence intervals. Here, Area under the ROC curve metric is used in subplot (a) and Average Precision in subplot (b).}\label{img:gbm_bench}
\end{figure}

It was hypothesized that LR might not be able to exploit the full potential of the input data (clinical variables and image assessments), as with this type of model, non-linear relationships within the data cannot be evaluated. Therefore, we utilized a GBM and trained it to predict the probability of OA progression. Figure \ref{img:gbm_bench} demonstrates the performance of models identical to model 1 and model 2, but trained using GBM instead of LR (model 3 and model 4). Model 4 performed best and obtained the AUC of 0.76 (0.75-0.78) and AP of 0.63 (0.61-0.65). The full comparisons of the models built using LR and GBM approaches are summarized in Table \ref{tab:model_benchmark} and also in Figures \ref{img:lr_bench} and \ref{img:gbm_bench}.

\begin{table}[ht!]
\centering
\caption{Summary of the reference models’ performances on the test set. Top performing models are underlined. 95\% confidence intervals are reported in parentheses. }\label{tab:model_benchmark}
\begin{tabular}{ccccc}
\toprule
\multirow{2}{*}{\textbf{Model}}                                                       & \multicolumn{2}{c}{\textbf{AUC}}                                                                                      & \multicolumn{2}{c}{\textbf{AP}}                                                                                    \T\B \\ \cline{2-5} 
                                                                                      & \textbf{LR}                                             & \textbf{GBM}                                                & \textbf{LR}                                           & \textbf{GBM}                                                \T\B  \\ 
\midrule
\begin{tabular}[c]{cc}\multirow{2}{*}{Age, Sex, BMI} & \\ & \end{tabular}                                                                         & 0.65 (0.63-0.67)                                        & 0.64 (0.63-0.66)                                            & 0.53 (0.51-0.55)                                      & 0.52 (0.49-0.54)                                            \\ \midrule
\begin{tabular}[c]{@{}c@{}}Age, Sex, BMI, Injury, \\ Surgery, WOMAC\end{tabular}   & 0.68 (0.66-0.69)                                        & 0.68 (0.66-0.69)                                            & 0.56 (0.53-0.58)                                      & 0.56 (0.53-0.58)                                            \\\midrule
\begin{tabular}[c]{cc}\multirow{2}{*}{KL-grade} & \\ & \end{tabular}                                                                              & 0.73 (0.71-0.75)                                        & -                                                           & 0.57 (0.55-0.58)                                      & -                                                           \\\midrule
\begin{tabular}[c]{c} Age, Sex, BMI, \\ KL-grade  \end{tabular}                                                                     & 0.75 (0.74-0.77)                                        & 0.76 (0.74-0.77)                                            & 0.61 (0.59-0.63)                                      & 0.61 (0.59-0.63)                                            \\\midrule
\begin{tabular}[c]{@{}c@{}}Age, Sex, BMI, Injury, \\ Surgery, WOMAC, KL-grade\end{tabular}                                       & 0.75 (0.74, 0.77)                                       & {\ul 0.76 (0.75-0.78)}                                      & 0.62 (0.60-0.64)                                      & {\ul 0.63 (0.61-0.65)}                                      \\ 
\bottomrule
\multicolumn{5}{l}{
\small
\begin{tabular}[c]{@{}l@{}}BMI -- Body-Mass Index\T \\
WOMAC -- Western Ontario and McMaster Universities Arthritis Index\\ 
KL-grade -- Kellgren-Lawrence grade\\ 
AUC -- Area Under the Receiver Operating Characteristic Curve\\ 
AP -- Average Precision\\ 
LR -- Logistic Regression\\ 
GBM -- Gradient Boosting Machine
\end{tabular}} 

\end{tabular}
\end{table}

\subsection*{Predicting progression from raw image data}
After testing the reference models, we developed a CNN, which allows to directly leverage raw knee DICOM images in an automatic manner. In contrast to the previous studies, this model was trained in a multi-task setting to predict OA progression in the index knee and also its current KL-grade from the corresponding X-ray image. In particular, our model consists of a feature extractor -- a pre-trained se-resnext50-32xd model \cite{hu2018squeeze} -- and two branches, each of which is a fully connected layer (FC), predicting its own task. One branch of the model predicts a progression outcome and the other branch a KL grade (Figure \ref{img:workflow}).

\begin{figure}[ht!]
 \centering
 \subfloat[]{%
   \includegraphics[width=0.48\textwidth]{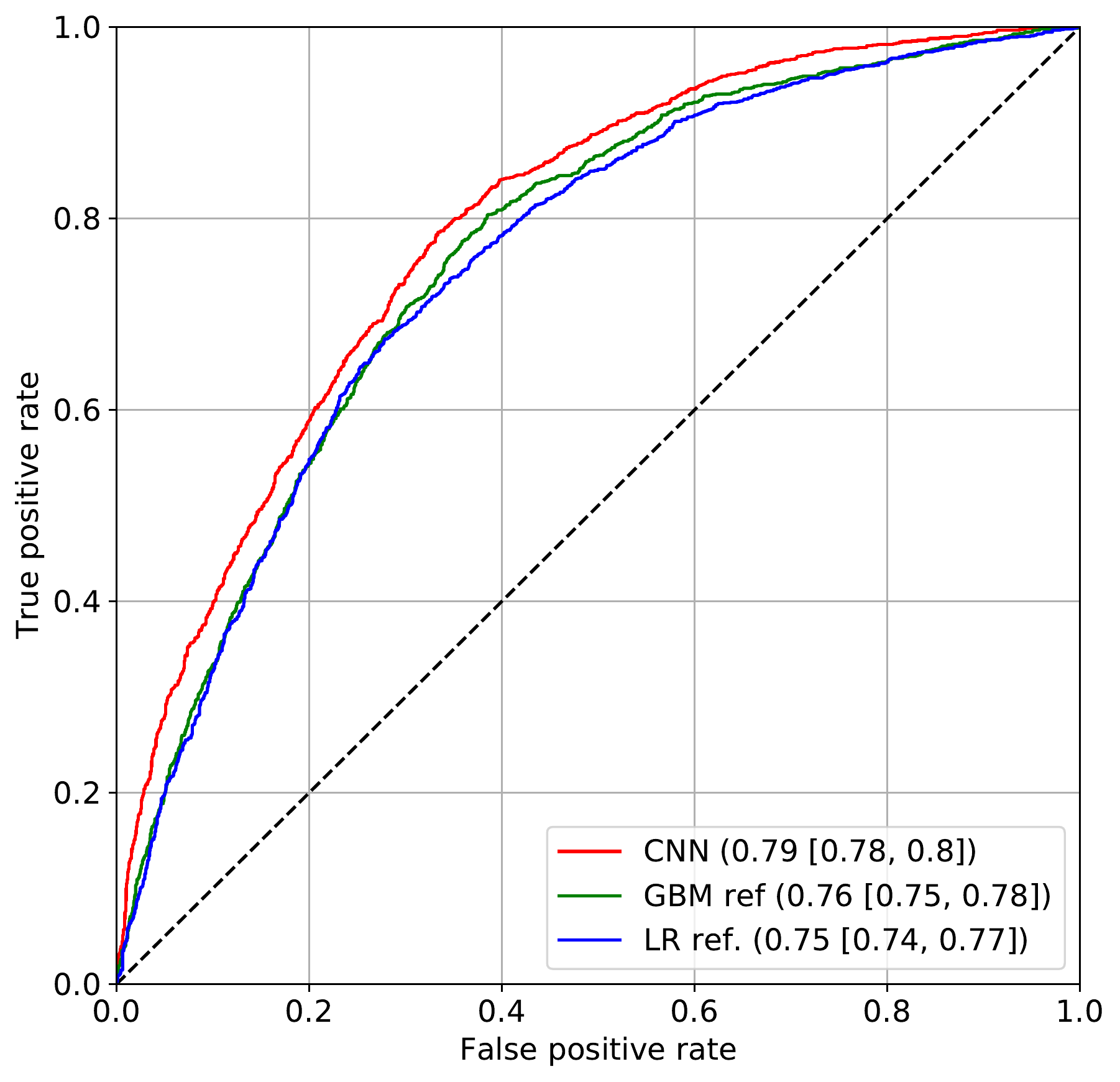}
   }
   \hfill
 \subfloat[]{%
   \includegraphics[width=0.48\textwidth]{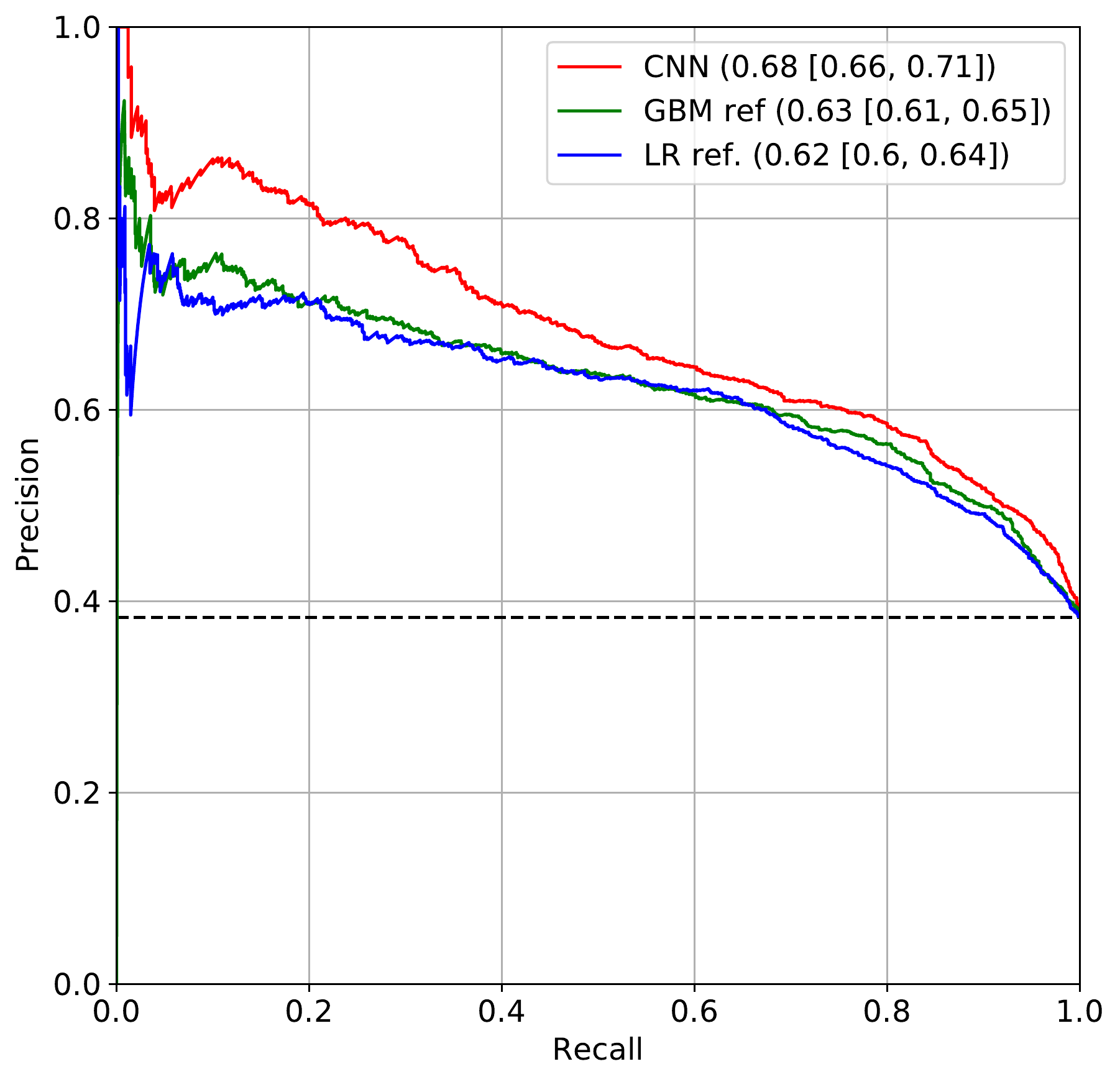}
   }
   
 \caption{Comparison of the deep convolutional neural network (CNN) and the reference methods built using Gradient Boosting Machine (GBM). Reference method based on Logistic Regression is also presented for better visual comparison (model 2 in the text). CNN model utilizes solely knee image and the GBM model utilizes KL grade and clinical data (model 4 in the text). Subplot (a) shows the ROC curves for CNN and GBM respectively. Subplot (b) shows the Precision-Recall Curves. Black dashed lines indicate the performance of a random classifier in case of AUC, and performance of the prediction model based on the dataset labels distribution. The subplots’ legends reflect the benchmarked models and the values of corresponding metrics with 95\% confidence intervals. Here, Area under the ROC curve metric is used in subplot (a) and Average Precision in subplot (b).}\label{img:cnn_bench}
\end{figure}

In our experiments, we found that prediction of the previously defined fine-grained classes -- no ($y=0$),  fast ($y=1$) and slow ($y=2$) progression, while being inaccurate individually, helps to regularize the training of the CNN and leads to better performance in predicting overall probability of progression $P(y>0 | x)$ within the following years. Having predicted such binary outcome, our CNN model (model 5) trained using the baseline knee image yielded AUC of 0.76 and AP of 0.56 in a cross-validation experiment on the training set. On the test set, the CNN yielded AUC of 0.79 (0.77-0.80) and AP of 0.68 (0.66-0.70). We compared this model to the strongest reference method -- model 4, and also strongest conventional method based on LR -- model 2 (Figure \ref{img:cnn_bench}). We obtained a statistically significant performance difference in AUC (DeLong’s $p$-value $< 1e-5$) when compared our CNN to the model 4. 

\begin{figure}[ht!]
\centering

 \subfloat[]{%
   \includegraphics[width=0.2\textwidth]{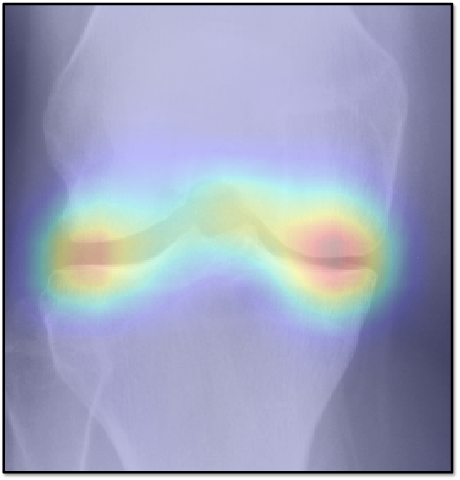}
 }
 \hfill
 \subfloat[]{%
    \includegraphics[width=0.2\textwidth]{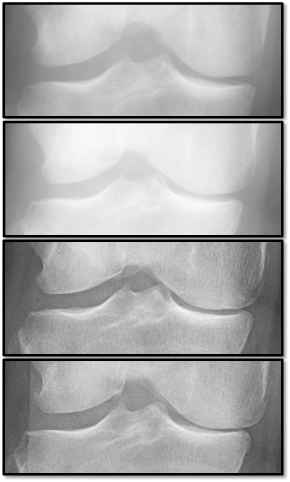}
 }
 \hfill
 \subfloat[]{%
   \includegraphics[width=0.2\textwidth]{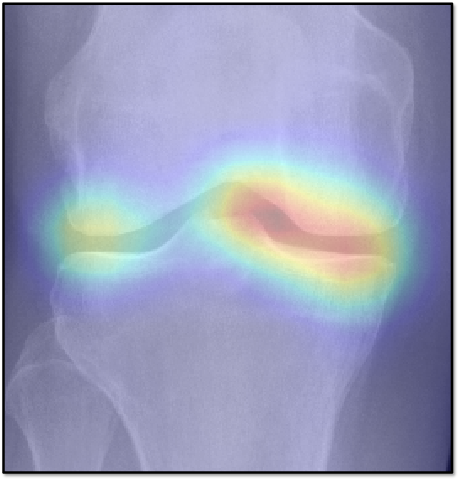}
 }
 \hfill
 \subfloat[]{%
   \includegraphics[width=0.2\textwidth]{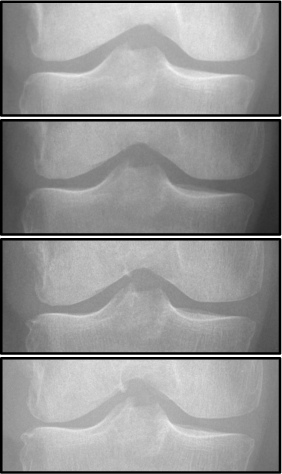}
  }

\caption{Examples of attention maps for progression cases and the corresponding visualization of progression derived using follow-up images from MOST datasets. Here, subplots (a) and (c) show the attention maps derived using a GradCAM approach. Subplots (b) and (d) show the joint-space areas from all the follow-up images (baseline to 84 months). Here, the subplot (b) corresponds to the attention map a) and the subplot (d) corresponds to the attention map (c).}\label{img:gcam_example}
\end{figure}

To gain insight into the basis of the CNN’s prediction, we used the GradCAM \cite{selvaraju2017grad} approach and visualized the attention maps for the well-predicted knees. Examples of attention maps are presented in Figure \ref{img:gcam_example}. We observed that in various cases, the CNN paid attention to the compartment opposite to the one where degenerative change became visible during the follow-up visits. Additional examples of such attention maps are presented in Supplementary Figures \ref{supp_img:gcam1}, \ref{supp_img:gcam2}, \ref{supp_img:gcam3} and \ref{supp_img:gcam4}.

To evaluate whether a combination of conventional diagnostic measures used in models 1-4 and CNN would further increase the predictive accuracy, we utilized a GBM in a stacked generalization fashion \cite{wolpert1992stacked} and treated both clinical measures and CNN's predictions as input features for the GBM (see Figure \ref{img:workflow}). Two stacked models were created. The first model, model 6, is fully automatic (does not use a KL-grade as an input) and predicts a probability of OA progression. It was built using all the predictions produced by the CNN -- $P(KL=i | x)$ for $i\in \{0,\dots,3\}$ and $P(y=i | x)$ for $i\in \{0,\dots, 2\}$, and additionally age, sex, BMI, knee injury history, knee surgery history and WOMAC total score.  The second model, model 7, was similar to the model 6, but with the addition of the KL grade that provides additional source information about the current stage of OA to the GBM. More details on building and training this two-stage pipeline are given in Methods. We hypothesized that a radiologist and a neural network may assign a KL grade differently, therefore, the difference in gradings could be leveraged for the prediction model, \textit{e.g.} if these gradings differ.

\begin{figure}[ht!]
 \centering
 \subfloat[]{%
   \includegraphics[width=0.48\textwidth]{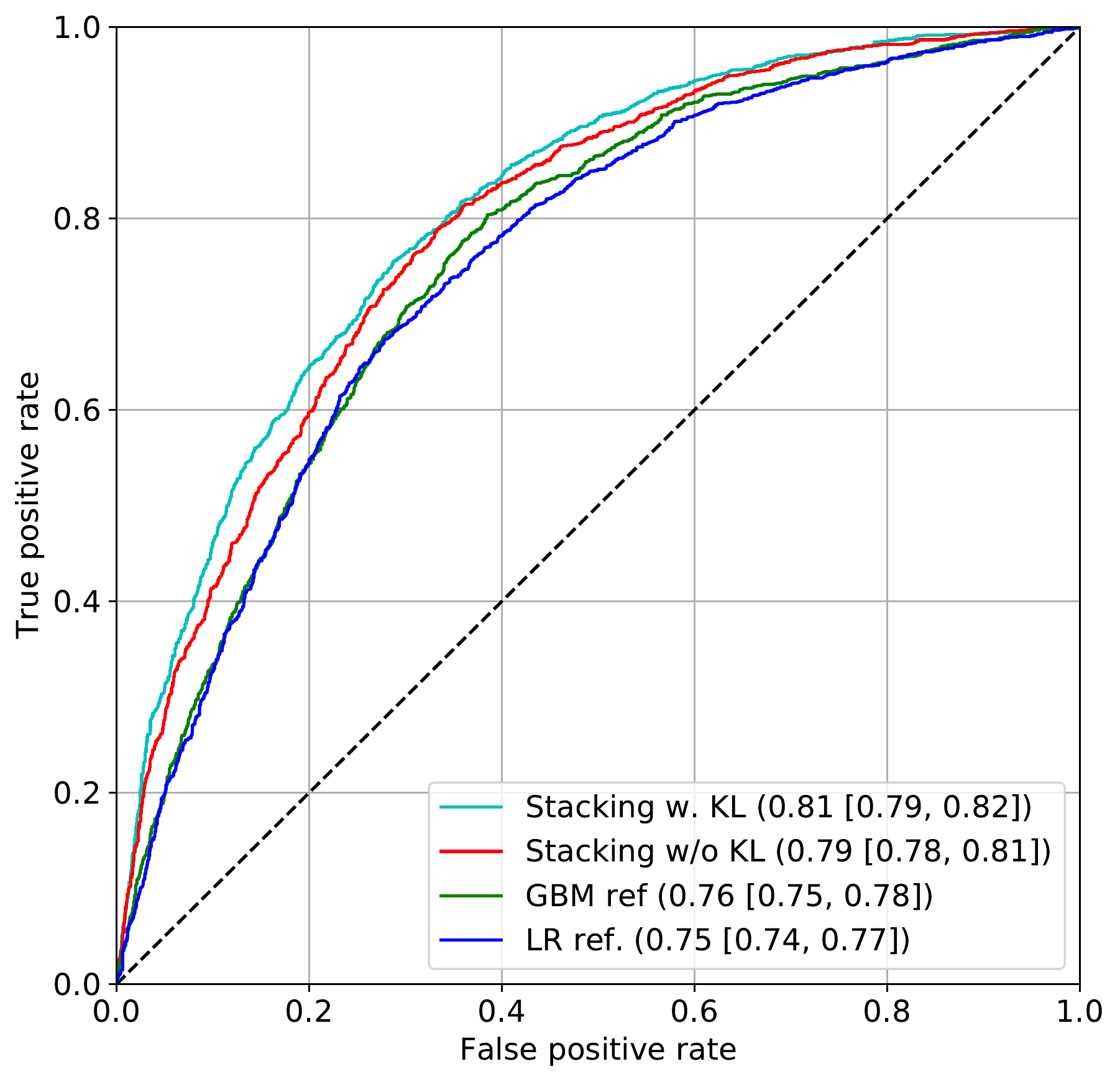}
   }
   \hfill
 \subfloat[]{%
   \includegraphics[width=0.48\textwidth]{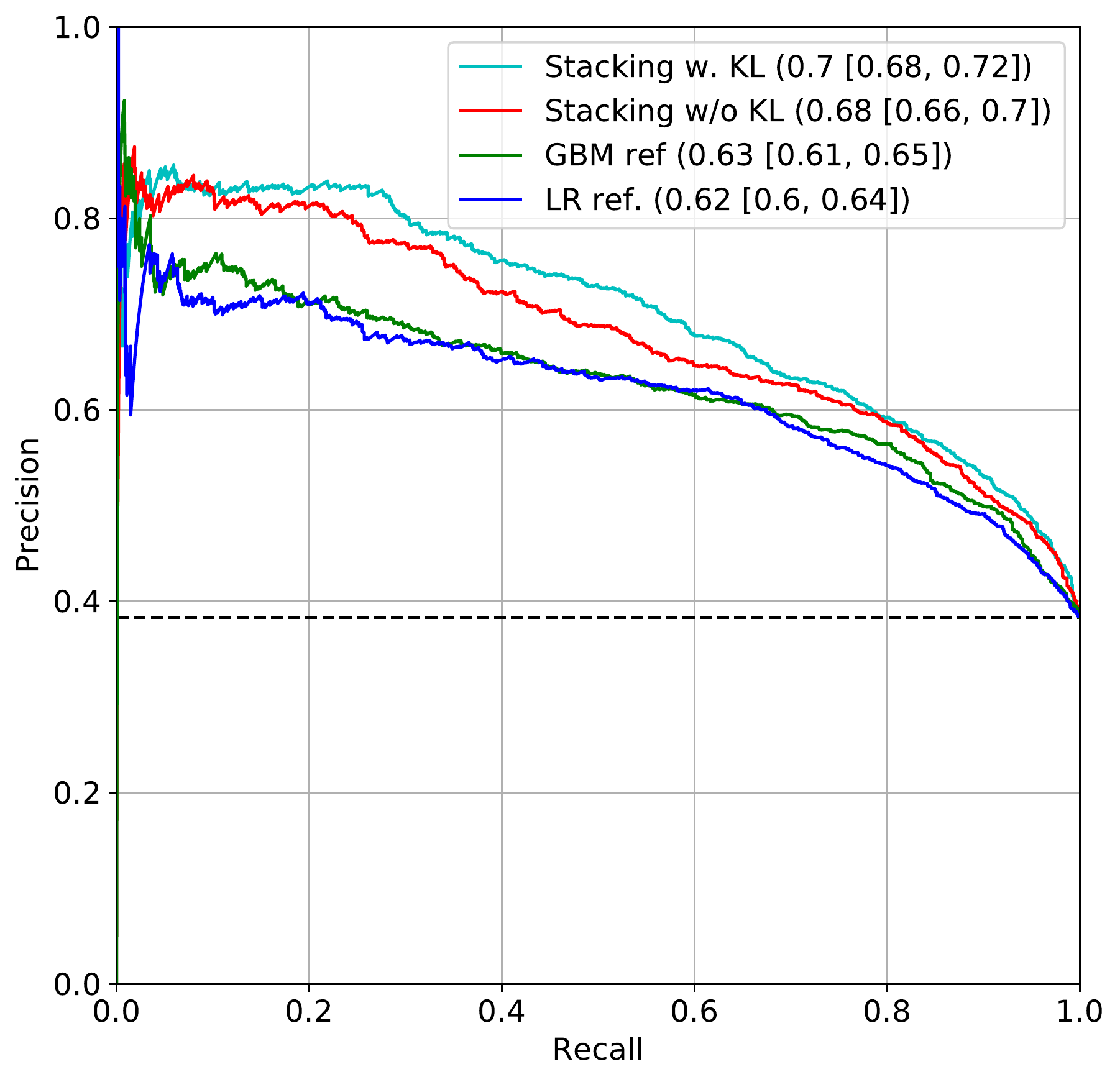}
   }
   
 \caption{Comparison of the multi-modal methods, based on Deep Convolutional Neural Network (CNN) and Gradient Boosting Machine (GBM) classifier versus the strongest reference method (model 4). Reference method based on Logistic Regression is also presented for better visual comparison (model 2). The subplots’ legends reflect the benchmarked models and the values of corresponding metrics with confidence intervals. Black dashed lines indicate the performance of a random classifier in case of AUC, and performance of the prediction model based on the dataset labels distribution. Here, Area under the ROC curve is used in subplot a) and Average Precision in subplot (b). The subplots (a) and (b) show the ROC and Precision-Recall (PR) curves respectively.  The results in this plot indicate that our method benefits from the utilization of a KL-grade.}\label{img:stacking_benchmarks}
\end{figure}

Figure \ref{img:stacking_benchmarks} shows the ROC and PR curves of models 6 and 7, along with the best reference method, model 4. As reported earlier, this reference model yielded AUC of 0.76 (0.75-0.78) and AP of 0.63 (0.61-0.65). In contrast, our multi-modal methods without and with utilization of a KL grade -- model 6 and model 7, yielded AUC of 0.79 (0.78-0.81), AP of 0.68 (0.66-0.71) and AUC of 0.80 (0.79-0.82), AP of 0.70 (0.68-0.72) respectively. Additionally, we also show the ROC and PR curves for model 2 in Figure \ref{img:stacking_benchmarks}. In Table \ref{tab:final_benchmark}, we present a detailed comparison of models 2, 4, 5, 6 and 7.

\begin{table}[ht!]
\centering
\caption{Detailed comparison of the developed models for all subjects included into testing conducted on the MOST dataset. 95\% confidence intervals are reported in parentheses for each of the reported metric.}\label{tab:final_benchmark}
\begin{tabular}{@{}cccc@{}}
\toprule
\textbf{Model \#}                                                  & \textbf{Model}                                                                                                                                                         & \textbf{AUC}                                                           & \textbf{AP}                                                           \T\B   \\ 
\midrule
2                                                                      & \begin{tabular}[c]{@{}c@{}}Age, Sex, BMI, Injury, \\ Surgery, WOMAC, KL-grade (LR)\end{tabular}                                                                        & 0.75 (0.74-0.77)                                                       & 0.62 (0.60-0.64)                                                       \\\midrule
4                                                                      & \begin{tabular}[c]{@{}c@{}}Age, Sex, BMI, Injury, \\ Surgery, WOMAC, KL-grade (GBM)\end{tabular}                                                                       & 0.76 (0.75-0.78)                                                       & 0.63 (0.61-0.65)                                                       \\\midrule
5                                                                      & \begin{tabular}[c]{cc} \multirow{2}{*}{CNN} & \\ & \end{tabular}                                                                                                                                                                    & 0.79 (0.77-0.80)                                                       & 0.68 (0.66-0.70)                                                       \\\midrule
6                                                                      & \begin{tabular}[c]{@{}c@{}}CNN + Age, Sex, BMI, Injury, \\ Surgery, WOMAC (GBM-based fusion)\end{tabular}                                                           & 0.79 (0.78-0.81)                                                       & 0.68 (0.66-0.71)                                                       \\\midrule
7                                                                      & \begin{tabular}[c]{@{}c@{}}CNN + Age, Sex, BMI, Injury, \\ Surgery, WOMAC, KL-grade (GBM-based fusion)\end{tabular}                                                 & {\ul 0.80 (0.79-0.82)}                                                 & {\ul 0.70 (0.68-0.72)}                                                 \\ 
\bottomrule
\multicolumn{4}{l}{\small
\begin{tabular}[c]{@{}l@{}}
KL-grade -- Kellgren-Lawrence grade \T \\
CNN -- Deep Convolutional Neural Network\\ 
BMI -- Body-Mass Index\\ 
WOMAC -- Western Ontario and McMaster Universities Arthritis Index\\ 
AUC -- Area Under the Receiver Operating Characteristic Curve\\ 
AP -- Average Precision\\ 
LR -- Logistic regression\\ 
GBM -- Gradient Boosting Machine
\end{tabular}
}

\end{tabular}
\end{table}

Finally, we also present the results on predicting OA progression for the subgroup of knees identified as KL-0 or KL-1 at baseline. These results are presented in Table \ref{tab:model_benchmark_kl01}. The results for this particular group of knees show that our method is capable of identifying knees that will progress to OA in a fully automatic manner with high performance -- our two best models, model 6 and model 7, yielded AUC of 0.78 (0.76-0.80) and 0.80 (0.78-0.82) respectively, and AP of 0.58 (0.55-0.62) and 0.62 (0.58-0.65) respectively.

\begin{table}[ht!]
\centering
\caption{Detailed comparison of the developed models for knees identified with Kellgren-Lawrence grade 0 or 1, which is considered as absence of osteoarthritis. The testing was done on the Multicenter Osteoarthritis Study dataset. 95\% confidence intervals are reported in parentheses for each of the reported metric.}\label{tab:model_benchmark_kl01}
\begin{tabular}{ccccc}
\toprule
\textbf{Model \#} & \textbf{Model} & \textbf{AUC} & \textbf{AP}  \T\B \\ \midrule
2 & \begin{tabular}[c]{@{}c@{}}Age, Sex, BMI, Injury, \\ Surgery, WOMAC, KL-grade (LR)\end{tabular} & 0.73 (0.70-0.75) & 0.52 (0.49-0.55) \\\midrule

4 & \begin{tabular}[c]{@{}c@{}}Age, Sex, BMI, Injury, \\ Surgery, WOMAC, KL-grade (GBM)\end{tabular} & 0.75 (0.72-0.77) & 0.54 (0.51-0.58) \\\midrule

5 & \begin{tabular}[c]{cc} \multirow{2}{*}{CNN} & \\ & \end{tabular} & 0.78 (0.76-0.80) & 0.58 (0.55-0.61) \\\midrule

6 & \begin{tabular}[c]{@{}c@{}}CNN + Age, Sex, BMI, Injury, \\ Surgery, WOMAC (GBM-based fusion)\end{tabular} & 0.78 (0.76-0.80) & 0.58 (0.55-0.62) \\\midrule

7 & \begin{tabular}[c]{@{}c@{}}CNN + Age, Sex, BMI, Injury, \\ Surgery, WOMAC, KL-grade (GBM-based fusion)\end{tabular} & {\ul 0.80 (0.78-0.82)} & {\ul 0.62 (0.58-0.65)} \\ \bottomrule
\multicolumn{4}{l}{
\small
\begin{tabular}[c]{@{}l@{}}KL-grade -- Kellgren-Lawrence grade \T \\ CNN -- Deep Convolutional Neural Network\\ BMI -- Body-Mass Index\\ WOMAC -- Western Ontario and McMaster Universities Arthritis Index\\ AUC -- Area Under the Receiver Operating Characteristic Curve\\ AP -- Average Precision\\ LR -- Logistic regression\\ GBM -- Gradient Boosting Machine\end{tabular}}
\end{tabular}
\end{table}
\section*{Discussion}
In this study, we presented a patient-specific machine learning-based method to predict structural knee OA progression from patient data acquired at a single clinical visit. The key difference of our method to the prior work is that it leverages the raw image of the patient's knee instead of any measures derived by human observers (e.g. JSW, KL or texture descriptors). 

The results presented in this study demonstrate that our method yields significantly better prediction performance than the conventionally used reference methods. The major finding of this study is that it is possible to predict knee OA progression from a single knee radiograph complemented with clinical data in a fully automatic manner. Other findings of this study demonstrate that the knee X-ray image alone is already a very powerful source of data to predict whether a particular knee will have OA progression or not. Finally, one of the main results from a clinical point of view is that it is possible to predict progression for patients having KL-0 and KL-1 at baseline. 

To the best of our knowledge, this is the first study where CNNs were utilized to predict OA progression directly from radiographs, and it is also one of the few studies in the field where an independent test set is used to robustly assess the results \cite{yu2019development, kerkhof2014prediction, hosnijeh2018development}. We believe that having such settings, where the test set remains unused until the final model’s validation, is crucial for further development of the OA progression prediction models. Another novelty of our approach is leveraging multi-modal patient data: plain radiographs (raw image data compared to KL-grades used previously \cite{kerkhof2014prediction, hosnijeh2018development} or manually designed texture parameters \cite{janvier2017subchondral, janvier2017subchondral2}), symptomatic assessment, and patient’s injury and/or surgery history data for prediction. Our results highlight that a combination of all the data allows to make more accurate predictions. Furthermore, thanks to GBM, with this approach it was possible to use missing data without imputation. 

In principle, clinical application of the developed method is straightforward and makes it possible to detect OA progression at a low cost in primary health care with minimal modifications to the current diagnostic chain. Our method can be utilized in a fully-automatic manner without a radiologist’s statement, and therefore, it could become available as an e.g. cloud service or software for physiotherapists to design behavioral interventions for the cases having high confidence of prediction. Compared to the other imaging modalities, such as MRI, the progression prediction methods developed just using radiographs and other easily obtainable data utilized in our study have potential to be the most accessible worldwide. 

While machine learning-based approaches yield stronger prediction than conventional statistical models, (\textit{e.g.} LR), they are less transparent, which can lead to lack of trust from clinicians. To address this drawback, various methods have been developed to explain the decisions of "black-box systems" \cite{selvaraju2017grad,olah2018building,bach2015pixel}. As such, we utilized the GradCAM approach \cite{selvaraju2017grad} that allowed us generating an attention map, in order to highlight the zones where the CNN has paid its attention. While being attractive, this approach can also lead to wrong interpretations, i.e. there is no theoretical guarantee that the neural network identifies causal relationships between image features and the output variable. Therefore, a thorough analysis of the attention maps is required to assess the significance of certain features and anatomical zones picked-up by the model. Such analysis, however, could enable new possibilities for investigation of the visual features. For example, we observed interesting associations in the GradCAM-generated attention maps (Figure \ref{img:gcam_example}), some of which are not captured by KL grading. As such, tibial spines (previously associated with OA progression \cite{kinds2013quantitative}) were highlighted in multiple attention maps. These associations, however, do not hold for all the progressors.

Although our study demonstrates a novel method, which outperforms various state-of-the art reference approaches, it also has several important limitations. Firstly, our model has not been tested in other populations than the ones from the United States. Testing the developed model on data from other populations would be a crucial step to bring the developed machine learning-based approach to primary healthcare. Secondly, we utilized only standardized radiographs  acquired with a positioning frame, which is not used in all the hospitals worldwide. Therefore, a validation of our model using the images acquired without the positioning frame is still needed. However, we tried to address this limitation by including data acquired under different beam angles to the test set. Thirdly, we relied only on the KL-grading system to define a progression outcome, and the symptomatic component of OA progression was completely ignored. This also needs to be addressed in the future studies. Finally, we used imputation in the test set when evaluating LR models. This could potentially lower the performance of LR-based reference methods. In contrast, GBM-based approach allowed us to leverage all the samples with missing data without imputation.

The results presented in this study show that, for subjects at risk, our proposed knee OA progression prediction model allows to identify the progressor cases on average 6\% more accurately than with the methods previously used in the OA literature. This study is an important step towards speeding up the OA disease modifying drug development process and also towards the development of better personalized treatment plans.

\section*{Methods}
\subsection*{Data description and pre-processing}
We utilized Osteoarthritis Initiative (OAI, \url{https://data-archive.nimh.nih.gov/oai}) and  Multicenter Osteoarthritis Study (MOST, \url{http://most.ucsf.edu}) follow-up cohorts. Both OAI and MOST datasets include clinical and imaging data from subjects at risk of developing OA 45-79 and 50-79 years old, from baseline to 96 (9 imaging follow-ups) and 84 months (4 imaging follow-ups), respectively. OAI dataset includes bilateral posterior-anterior knee images, acquired with a Synaflexer\texttrademark frame \cite{kothari2004fixed} and 10 degrees beam angle, while the MOST dataset also has images acquired with 5- and 15-degrees beam angles. 

Our inclusion criteria were the following. Firstly, we excluded the knees that had TKA, end-stage OA (KL-4) or had a missing KL-data at the baseline. Subsequently, we excluded the knees which did not progress and were not examined at the last follow-up. This allowed us to ensure that the subjects in the train and test sets did not progress within 96 and 84 months, respectively. If the knee had any increase of the KL-grade during the follow-up, we assigned the class of the earliest noticed KL-grade increase, e.g. if the knee progressed at 30 months and 84 months, we used 30-months follow-up visit to define the fine-grained progression class. Data selection flowcharts for OAI and MOST datasets are presented in Supplementary Figures \ref{supp_img:oai_flowchart} and \ref{supp_img:most_flowchart}, respectively. The exact implementation of this selection process is also presented in the supplied source code (see Data Availability Statement).
	
In our experiments, we utilized variables such as age, sex, BMI, injury history, surgery history and total WOMAC (Western Ontario and McMaster Universities Arthritis Index) score. Due to the presence of missing values, it would be impossible to train and test LR model without utilizing imputation techniques or removing the missing data. Therefore, during the training of LR, we excluded the knees with missing values. In the test dataset (MOST), we imputed the missing variables by utilizing mean value imputation strategy when testing the LR. When we trained GBM-based method, the imputation strategies are not needed, thus we used the data extracted from MOST metadata as is.

\subsection*{Image pre-processing}
To pre-process the OAI and MOST DICOM images, for each knee we extracted a region of interest (ROI) of $140\times140$ mm using an ad-hoc script and BoneFinder software \cite{lindner2015robust} that enables accurate landmark localization using regression voting approach. This was done in order to standardize the coordinate frame among the patients and the data acquisition centers. After localizing the bone landmarks, we rotated all the knee images so that the tibial plateau was horizontal. Subsequently, we performed a histogram clipping between $5^{th}$ and $99^{th}$ percentiles and used global contrast normalization subtracting the image minimum and dividing all the image pixels by the maximum pixel value. Then, we converted the images to 8-bit depth multiplying them by 255. Finally, all the images were resized to $310\times310$ pixels (new pixel spacing of 0.45 mm) and the left knee images were flipped horizontally to match the collateral (right) knee. 

\subsection*{Experimental setup and reference methods}
All experiments, including the hyper-parameter search, were carried out using the same 5-fold subject-wise cross-validation on OAI data. A stratified cross-validation was used to obtain the same distribution of progressed and non-progressed cases in both train and validation splits for each fold. To implement this validation scheme, we used the publicly available scikit-learn package \cite{pedregosa2011scikit}. 

For building regularized LR models, we used scikit-learn and for non-regularized LR we used the statsmodels package \cite{seabold2010statsmodels}. For GBM models, we utilized the LightGBM \cite{ke2017lightgbm} implementation. We built the CNN models using PyTorch 1.0 \cite{paszke2017automatic} and trained them using three NVidia GTX 1080Ti cards.

To find the best hyperparameters set for GBM, we used the Bayesian hyperparameters optimization package hyperopt \cite{bergstra2013hyperopt} with 500 trials. Each trial maximized the AP on cross-validation. In the case of CNN, we also used cross-validation and built 5 models. We used the snapshot of the model’s weights that yielded the maximum AP value on the validation set in each cross-validation split. The hyperparameters for CNN were found empirically.

\subsection*{Deep neural network’s implementation details}
We designed a multi-task CNN architecture to predict OA progression, and our model consisted of a convolutional (Conv) and two fully-connected (FC) blocks. One FC layer had three outputs corresponding to the three progression classes, and the other had 5 outputs, corresponding to the prediction of the current -- baseline KL grade. This is schematically illustrated in Figure 1. To harmonize the size of the outputs after Conv layers and the inputs of the FC layers, we utilized a Global Average Pooling layer.

We used the design of the Conv layers from se-resnext50\_32x4d network \cite{hu2018squeeze}. In the initial cross-validation experiments, we also evaluated se-resnet50, inceptionv4, se-resnext101\_32x4d; however, we did not obtain significantly better results than the ones reported in this study. To train the CNN, we utilized a transfer learning similarly to \cite{tiulpin2018automatic} and initialized the weights of all the Conv layers from a network trained on the ImageNet dataset \cite{deng2009imagenet}. The two FC layers were initialized from random noise. 

In contrast to the FC layers, the weights of the Conv layers were not trained during the first 2 epochs (full passes through the training set) and then they were unfrozen. Subsequently, all the layers of the CNN were trained for 20 epochs. Such strategy ensured that the FC layers did not corrupt the pre-trained Conv weights during the first backpropagation passes. The CNN was trained with a learning rate of $1e-3$ (dropped at $15^{th}$ epoch), batch size of 64, weight decay of $1e-4$ and Adam optimization method \cite{kingma2014adam}. We also placed a dropout layer \cite{srivastava2014dropout} with the rate of $p=0.5$ before each FC layer.

During the training of the CNN, we used random noise addition, random rotation $\pm 5$ degrees, random cropping of the original $310\times 310$ pixels image to $300\times 300$ pixels ($135\times 135$ mm) and also random gamma correction. These data augmentations were performed randomly on-the-fly, with the aim to train our model to be invariant towards different data acquisition parameters. We used the SOLT package of version 0.1.3 \cite{tiulpin2019solt} in our experiments.

\subsection*{Inference pipeline}
At the test phase, we averaged the outputs of all the models trained in cross-validation. Additionally, for each CNN model here, we performed 5-crop test-time augmentation (TTA). Specifically, we cropped 4 images of $300\times 300$ pixels from the corners of the original image, and one same-sized crop from the center of the image. The predictions for the 5 cropped images were eventually averaged. Subsequently, having the TTA prediction for each cross-validation model, we averaged their results as well. This approach allowed us to reduce the variance of the CNNs and boost the prediction accuracy. 

It is worth to mention that during the evaluation of CNN model alone, instead of using the fine-grained division into progression classes, we used the probability of progression $P(prog | x)$ as a sum of $P(y=1 | x)$ and $P(y=2 | x)$. A similar technique was previously utilized in a skin cancer prediction study \cite{esteva2017dermatologist}.  

\subsection*{Interpreting neural network’s decisions}
In this study, we focused not only on producing the first state-of-the-art model for knee OA progression prediction, but also developed an approach to examine the network’s decision to assess the radiological features detected by the network. Similar to our previous study \cite{tiulpin2018automatic}, we modified the GradCAM method \cite{selvaraju2017grad} to operate with TTA. The output of the GradCAM is an attention map, showing which region of the image positively correlates with the output of the network.

In the previous section, we described a TTA-approach and it should be noted that all the operations including the sum of the progression probabilities are fully differentiable, thus the application of the GradCAM here is fairly straightforward. 

\subsection*{Model stacking: fusing heterogeneous data using tree gradient boosting} 
We fused the predictions of the neural network -- KL grade and progression probabilities $P(KL=i | x)$, $i\in \{0, \dots, 4\}$ and $P(y=i|x)$, $i\in \{0, 1, 2\}$ respectively -- with other clinical measures such as patient’s age, sex, BMI, previous injury history, symptomatic assessments (WOMAC) and, optionally, a KL grade. Such fusion is challenging, prone to overfitting and requires a robust cross-validation scheme. A stacked generalization approach, proposed by Wolpert \cite{wolpert1992stacked} allows to build multiple layers of models and handle these issues.

Following our model inference strategy, we first trained the 5 CNN models corresponding to the 5 cross-validation train-validation splits. Subsequently, this allowed to perform the inference on each validation set in our cross-validation setup and, therefore, obtain CNN predictions for the whole training set. When building the second-level GBM, we utilized the same cross-validation split and used the predictions for each knee joint as input features, along with the other clinical measures.

\subsection*{Statistical analyses}
We utilized Precision-Recall (PR) and ROC curves as the main methods to measure the performance of all the methods. PR curve can be quantitatively summarized using the AP metric. The AP metric gives a general understanding on average positive predictive value (PPV) of the method. PPV indicates the probability of the object predicted as positive (progressor in the case of this study) actually being positive. The precision-recall curve has been shown to be more informative than the ROC curve when comparing classifiers on imbalanced datasets \cite{saito2015precision}. ROC curve can quantitatively be summarized using the AUC. ROC curve demonstrates a trade-off between the true positive rate (sensitivity) and the false positive rate (1 - specificity) of the classifier. AUC represents the quality of ranking random positive examples over the random negative examples \cite{cortes2004auc}.
	
To compute the AUC and AP on the test set, we used stratified bootstrapping with 2,000 iterations. The stratification allowed us to reliably assess the confidence intervals for both AUC and AP. We assessed the statistical significance of the difference between the models using DeLong’s test \cite{delong1988comparing}. 

\subsection*{Data Availability Statement}
OAI and MOST datasets are publicly available datasets and can be requested at http://most.ucsf.edu/ and https://oai.epi-ucsf.org/. The Dockerfile, source codes, pre-trained models and other relevant data are publicly available (https://github.com/MIPT-Oulu/OAProgression).

\subsection*{Acknowledgements}
The OAI is a public-private partnership comprised of five contracts (N01- AR-2-2258; N01-AR-2-2259; N01-AR-2- 2260; N01-AR-2-2261; N01-AR-2-2262) funded by the National Institutes of Health, a branch of the Department of Health and Human Services, and conducted by the OAI Study Investigators. Private funding partners include Merck Research Laboratories; Novartis Pharmaceuticals Corporation, GlaxoSmithKline; and Pfizer, Inc. Private sector funding for the OAI is managed by the Foundation for the National Institutes of Health. 

MOST is comprised of four cooperative grants (Felson - AG18820; Torner - AG18832; Lewis - AG18947; and Nevitt - AG19069) funded by the National Institutes of Health, a branch of the Department of Health and Human Services, and conducted by MOST study investigators. This manuscript was prepared using MOST data and does not necessarily reflect the opinions or views of MOST investigators. 

We would like to acknowledge the strategic funding of the University of Oulu, Infotech Oulu, KAUTE foundation and Sigrid Juselius Foundation for supporting this work.

Dr. Claudia Lindner is acknowledged for providing BoneFinder.

\section*{Author contributions}
A.T. and S.S. originated the idea of the study. 
A.T., S.S., and S.K. designed the study,
A.T. performed the experiments and wrote the manuscript
S.K., J.T., E.R. provided the technical feedback.
S.B., E.O. and J.M. provided the clinical feedback.
All authors participated in the manuscript writing and editing.

\section*{Additional information}
\subsection*{Competing financial interests}
The authors declare no competing financial interests.
\newpage
\bibliography{natcomm_oa_prog}
\newpage
\section*{Supplementary data}
\setcounter{figure}{0}  
\setcounter{table}{0}   

\begin{table}[!ht]
\centering
\caption{Subject-level characteristics for subsets of Osteoarthritis Initiative (OAI) and Multicenter Osteoarthritis Study (MOST) datasets, used in this study as train and test sets respectively.}
\label{supp_tab:subjects}
\begin{tabular}{ccccc}
\toprule
\textbf{Dataset} & \textbf{Age} & \textbf{BMI} & \textbf{\# Females} & \textbf{\# Males}  \T\B  \\
\midrule
OAI (Train)      & 61.16\rpm 9.19 & 28.62\rpm4.84 & 1,552                & 1,159              \\ \midrule
MOST (Test)      & 62.50\rpm 8.11 & 30.74\rpm5.97 & 1,303                & 826               \\
\bottomrule
\multicolumn{5}{l}{BMI – Body Mass Index}                                                \\ 
\end{tabular}
\end{table}
\begin{table}[!ht]
\caption{Knee-level characteristics for subsets of Osteoarthritis Initiative (OAI) and Multicenter Osteoarthritis Study (MOST) datasets, used in this study as train and test sets respectively. KL-0 to KL4 represent Kellgren-Lawrence Grading scale of osteoarthritis (OA) – from healthy knee to end-stage OA. Here, (P) indicates the knees, which progressed during the follow-up visits and (NP) the ones which did not progress.}
\centering
\label{supp_tab:knees}
\begin{tabular}{@{}cccccccccc@{}}
\toprule
\multirow{2}{*}{\textbf{Dataset}} & \multirow{2}{*}{\textbf{Subset}} & \multicolumn{5}{c}{\textbf{KL-grade}} & \multirow{2}{*}{\textbf{Total}} & \multirow{2}{*}{\textbf{\# Left}} & \multirow{2}{*}{\textbf{\# Right}}  \T\B \\ \cmidrule(lr){3-7}
                         &                         & 0      & 1   & 2   & 3   & 4 &                                 &                                   &                                    \\ \midrule
\multirow{2}{*}{OAI}     & NP                      & 2,133  & 702 & 569 & 193 & 0 & 3,597                           & 1,803                             & 1,794                              \\ \cmidrule(lr){2-10}
                         & P                       & 271    & 466 & 346 & 248 & 0 & 1,331                           & 654                               & 677                                \\ \midrule
\multirow{2}{*}{MOST}    & NP                      & 1,558  & 336 & 314 & 209 & 0 & 2,417                           & 1,208                             & 1,209                              \\ \cmidrule(lr){2-10}
                         & NP                      & 322    & 387 & 380 & 412 & 0 & 1,501                           & 716                               & 785                                \\ 
\bottomrule
\end{tabular}
\end{table}

\begin{figure}[ht!]
\centering

\includegraphics[width=\textwidth]{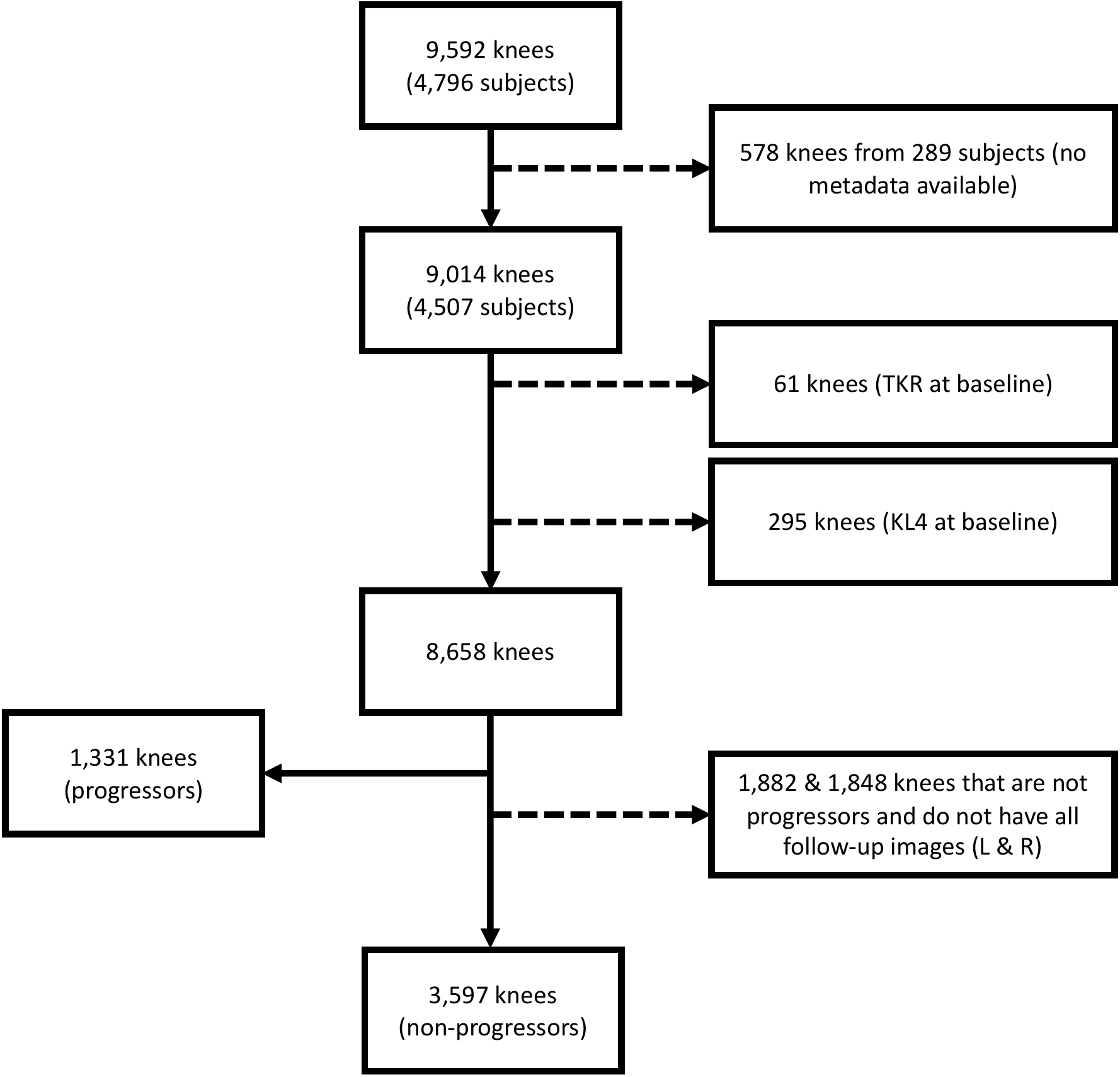}
\caption{Data selection flowchart for Osteoarthritis Initiative (OAI) dataset which was used to train the model.}\label{supp_img:oai_flowchart}
\end{figure}

\begin{figure}[!ht]
\centering

\includegraphics[width=0.85\textwidth]{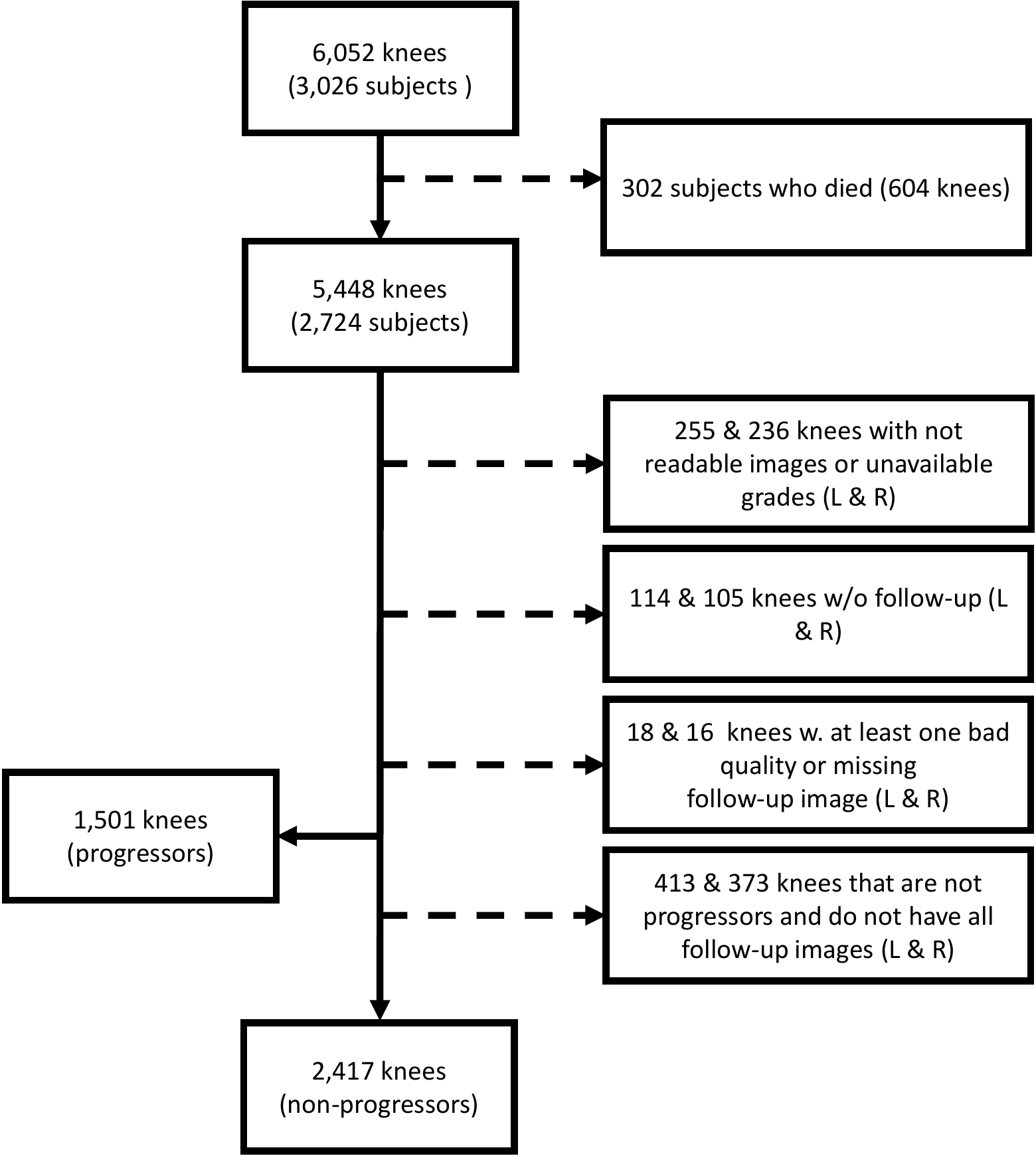}
\caption{Data selection flowchart for Multicenter Osteoarthritis Study (MOST) dataset which was used to test the model.}\label{supp_img:most_flowchart}
\end{figure}

\begin{figure}[!ht]
\centering

\null\hfill
\subfloat[KL-0 to KL-2, slow]{\includegraphics[width=0.35\textwidth]{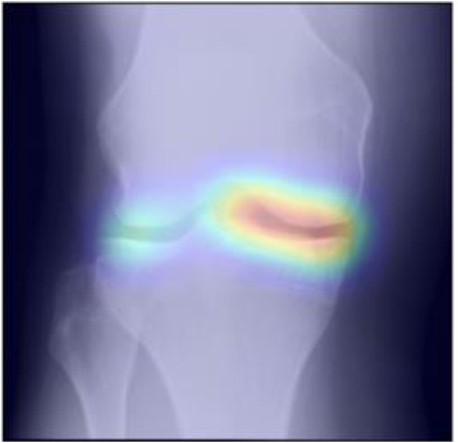}}
\hfill
\subfloat[KL-0 to KL-3, slow]{\includegraphics[width=0.35\textwidth]{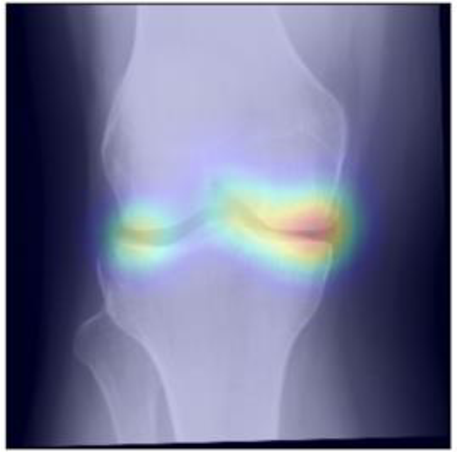}}
\hfill\null

\null\hfill
\subfloat[KL-0 to KL2, slow]{\includegraphics[width=0.35\textwidth]{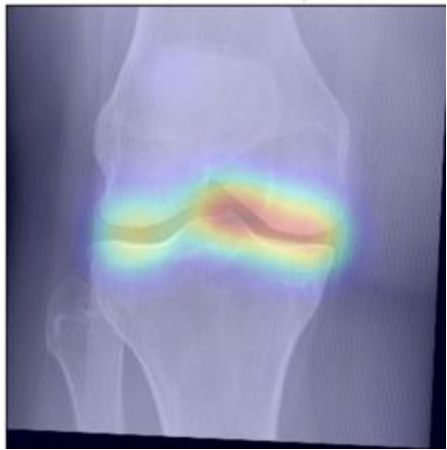}}
\hfill
\subfloat[KL-1 to KL-3, slow]{\includegraphics[width=0.35\textwidth]{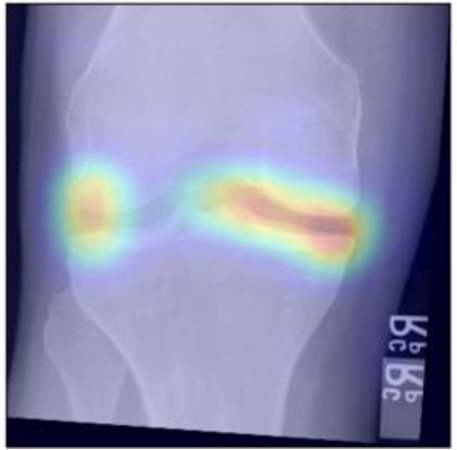}}
\hfill\null

\null\hfill
\subfloat[KL-1 to KL-2, fast]{\includegraphics[width=0.35\textwidth]{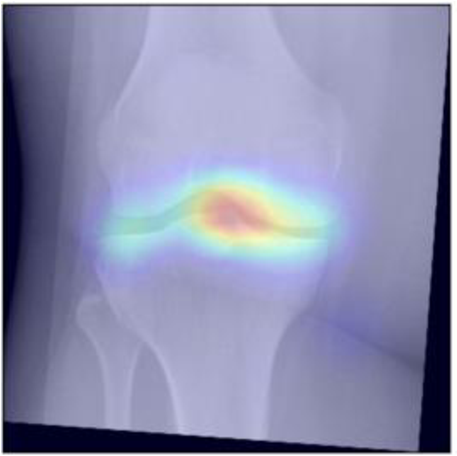}}
\hfill
\subfloat[KL-1 to KL3, fast]{\includegraphics[width=0.35\textwidth]{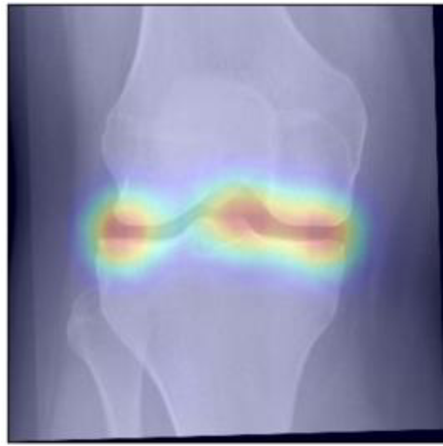}}
\hfill\null

\caption{Examples of GradCAM-based attention maps for the knees progressed from no osteoarthritis to osteoarthritis. Fine-grained sub-types of progression are also specified. The presented images are of $140\times 140$ mm.}\label{supp_img:gcam1}
\end{figure}

\begin{figure}[!ht]
\centering
\null\hfill
\subfloat[KL-2 to KL-3, slow]{\includegraphics[width=0.35\textwidth]{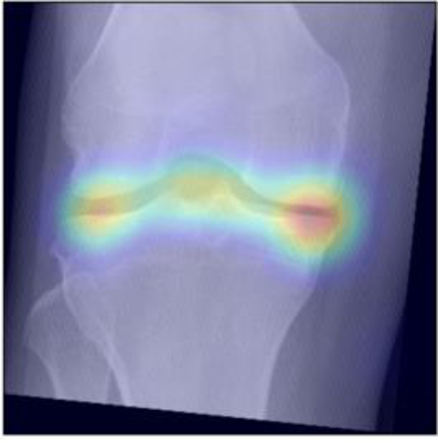}}
\hfill
\subfloat[KL-2 to KL-3, fast]{\includegraphics[width=0.35\textwidth]{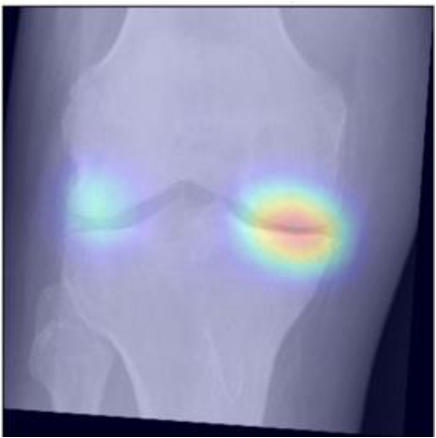}}
\hfill\null
 
\null\hfill
\subfloat[KL-3 to KL-4, slow]{\includegraphics[width=0.35\textwidth]{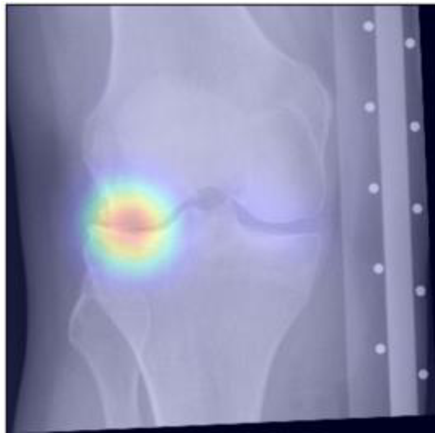}}
\hfill
\subfloat[KL-3 to TKR]{\includegraphics[width=0.35\textwidth]{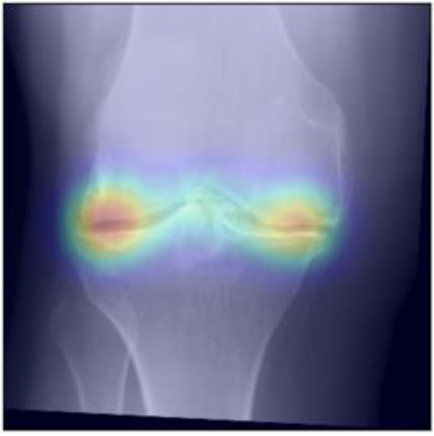}}
\hfill\null

\null\hfill
\subfloat[KL-2 to KL-3, fast]{\includegraphics[width=0.35\textwidth]{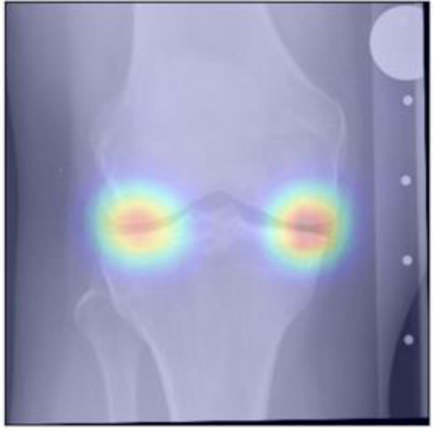}}
\hfill
\subfloat[KL-3 to KL-4, fast]{\includegraphics[width=0.35\textwidth]{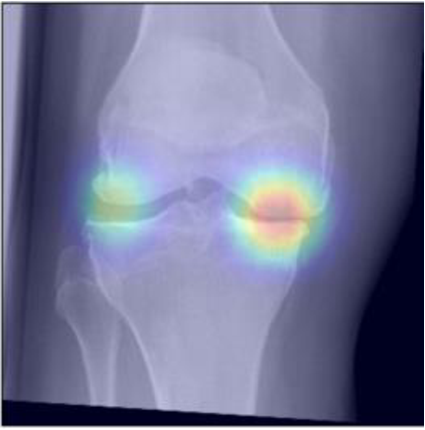}}
\hfill\null

\caption{Examples of GradCAM-based attention maps for the knees having osteoarthritis at baseline and progressed in the future. Fine-grained sub-types of progression are also specified. The presented images are of $140\times 140$ mm.}\label{supp_img:gcam2}
\end{figure}

\begin{figure}[!ht]
\centering
\null\hfill
\subfloat[KL-1]{\includegraphics[width=0.35\textwidth]{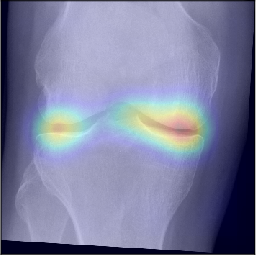}}
\hfill
\subfloat[KL-0]{\includegraphics[width=0.35\textwidth]{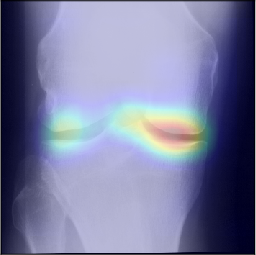}}
\hfill\null

\null\hfill
\subfloat[KL-1]{\includegraphics[width=0.35\textwidth]{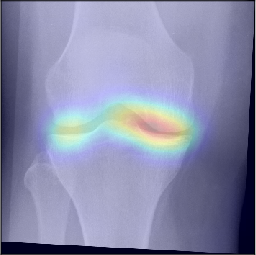}}
\hfill
\subfloat[KL-0]{\includegraphics[width=0.35\textwidth]{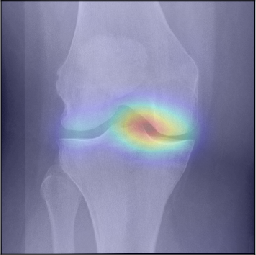}}
\hfill\null

\null\hfill
\subfloat[KL-1]{\includegraphics[width=0.35\textwidth]{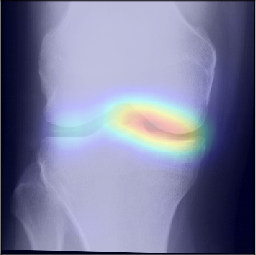}}
\hfill
\subfloat[KL-1]{\includegraphics[width=0.35\textwidth]{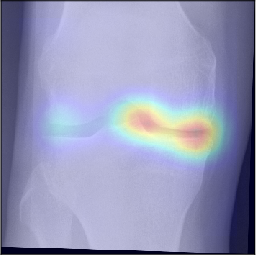}}
\hfill\null

\caption{Examples of GradCAM-based attention maps for the knees having no osteoarthritis at baseline and that did progress within the next 7 years. Baseline Kellgren-Lawrence (KL) grades are specified. The presented images are of $140\times 140$ mm.}\label{supp_img:gcam3}
\end{figure}

\begin{figure}[!ht]
\centering
\null\hfill
\subfloat[KL-2]{\includegraphics[width=0.35\textwidth]{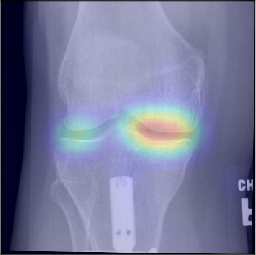}}
\hfill
\subfloat[KL-2]{\includegraphics[width=0.35\textwidth]{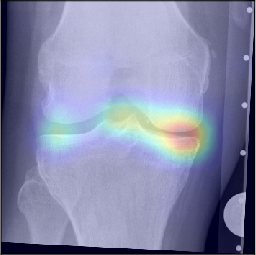}}
\hfill\null

\null\hfill
\subfloat[KL-2]{\includegraphics[width=0.35\textwidth]{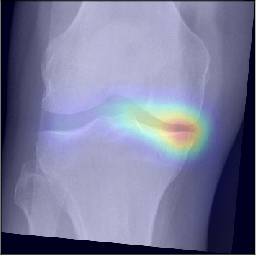}}
\hfill
\subfloat[KL-2]{\includegraphics[width=0.35\textwidth]{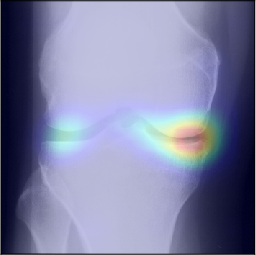}}
\hfill\null

\null\hfill
\subfloat[KL-2]{\includegraphics[width=0.35\textwidth]{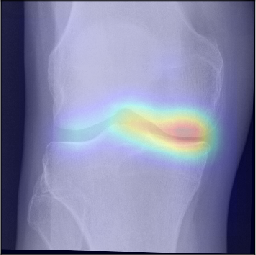}}
\hfill
\subfloat[KL-2]{\includegraphics[width=0.35\textwidth]{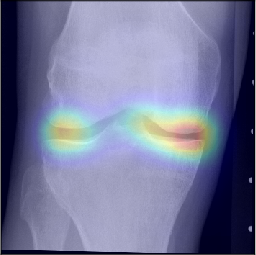}}
\hfill\null

\caption{Examples of GradCAM-based attention maps for the knees having early osteoarthritis at the baseline and that did not progress withing the next 7 years. Baseline Kellgren-Lawrence (KL) grades are specified. The presented images are of $140\times 140$ mm.}\label{supp_img:gcam4}
\end{figure}

\end{document}